\def\year{2021}\relax
\documentclass[letterpaper]{article} 
\usepackage{aaai21}  
\usepackage{times}  
\usepackage{helvet} 
\usepackage{courier}  
\usepackage[hyphens]{url}  
\usepackage{graphicx} 
\usepackage{natbib}  
\usepackage{caption} 
\usepackage{amsmath,amssymb} 
\usepackage{algorithm}
\usepackage{algorithmic}
\usepackage{booktabs}
\title{Selective Pseudo-Labeling with Reinforcement Learning for Semi-Supervised Domain Adaptation}
\author{
    Bingyu Liu\textsuperscript{\rm 1, \rm 3}, 
    Yuhong Guo\textsuperscript{\rm 1, \rm 2}, 
    Jieping Ye\textsuperscript{\rm 1}, 
    Weihong Deng\textsuperscript{\rm 3}\\
}
\affiliations{
    \textsuperscript{\rm 1} Didi Chuxing\quad
    \textsuperscript{\rm 2} Carleton University\quad
    \textsuperscript{\rm 3} Beijing University of Posts and Telecommunications
}
\begin{document}

\maketitle

\begin{abstract}
Recent domain adaptation methods have demonstrated impressive improvement on unsupervised domain adaptation problems. However, in the semi-supervised domain adaptation (SSDA) setting where the target domain has a few labeled instances available, these methods can fail to improve performance. Inspired by the effectiveness of pseudo-labels in domain adaptation, we propose a reinforcement learning based selective pseudo-labeling method for semi-supervised domain adaptation. It is difficult for conventional pseudo-labeling methods to balance the correctness and representativeness of pseudo-labeled data. To address this limitation, we develop a deep Q-learning model to select both accurate and representative pseudo-labeled instances. Moreover, motivated by large margin loss's capacity on learning discriminative features with little data, we further propose a novel target margin loss for our base model training to improve its discriminability. Our proposed method is evaluated on several benchmark datasets for SSDA, and demonstrates superior performance to all the comparison methods.
\end{abstract}

\section{Introduction}

Deep convolutional neural networks (CNNs)~\cite{simonyan2014very,krizhevsky2012imagenet,he2016deep,szegedy2015going,xie2017aggregated,wang2018deep,li2018deep} have achieved remarkable success in image classification tasks. When trained on large-scale labeled data, deep networks can learn discriminative representations and present great performance. However, it is difficult to collect and annotate datasets for many domains. A good option is to use the labeled data available in other domains for training models, which however often presents a domain shift challenge between the two domains and degrades the test performance. To address this problem, many unsupervised domain adaptation (UDA) methods~\cite{wang2018deepda,ganin2014unsupervised,long2015learning,long2018conditional,saito2017adversarial} have been proposed. UDA aims to improve the generalization performance on unlabeled target domains. However, in reality a few labeled target instances can be available in target domains, and this semi-supervised domain adaptation (SSDA) setting is more common. According to~\cite{saito2019semi}, UDA methods often fail to improve performance compared with just training on the unified data of labeled source and target samples in the semi-supervised setting.

For the task of domain adaptation, the purpose is to improve the generalization performance in the target domain. In the SSDA setting, we have a few labeled target samples, but the number of them is too small to represent the distribution of target unlabeled data. To increase the number of labeled instances in the target domain without incurring annotation cost, one intuitive strategy is to exploit pseudo-labels of the target domain samples produced by a current prediction model. However, the pseudo-labels can often be very noisy and contain many wrong labels, while training with the mislabeled samples can negatively impact the original model. This motivates the straightforward selective pseudo-labeling strategy which selects the most confident pseudo-labels to increase their probability of correctness~\cite{chen2019progressive}. This simple strategy can select more accurate samples but non-necessary the most useful samples for the prediction model. For example, the more confident and accurate samples may be closer to the labeled data and fail to represent the distribution of the target domain. It is more reasonable but difficult to perform selective pseudo-labeling by balancing the accuracy and the representativeness of the selected samples. To address this challenge, in this paper we propose a reinforcement learning based selective pseudo-labeling method. Our strategy is to use deep Q-learning to learn appropriate selection policies with reward functions that reflect both factors of label correctness and data representativeness.

\begin{figure}[htbp]
\vskip .1in 
\begin{center}
   \includegraphics[width=0.98\columnwidth]{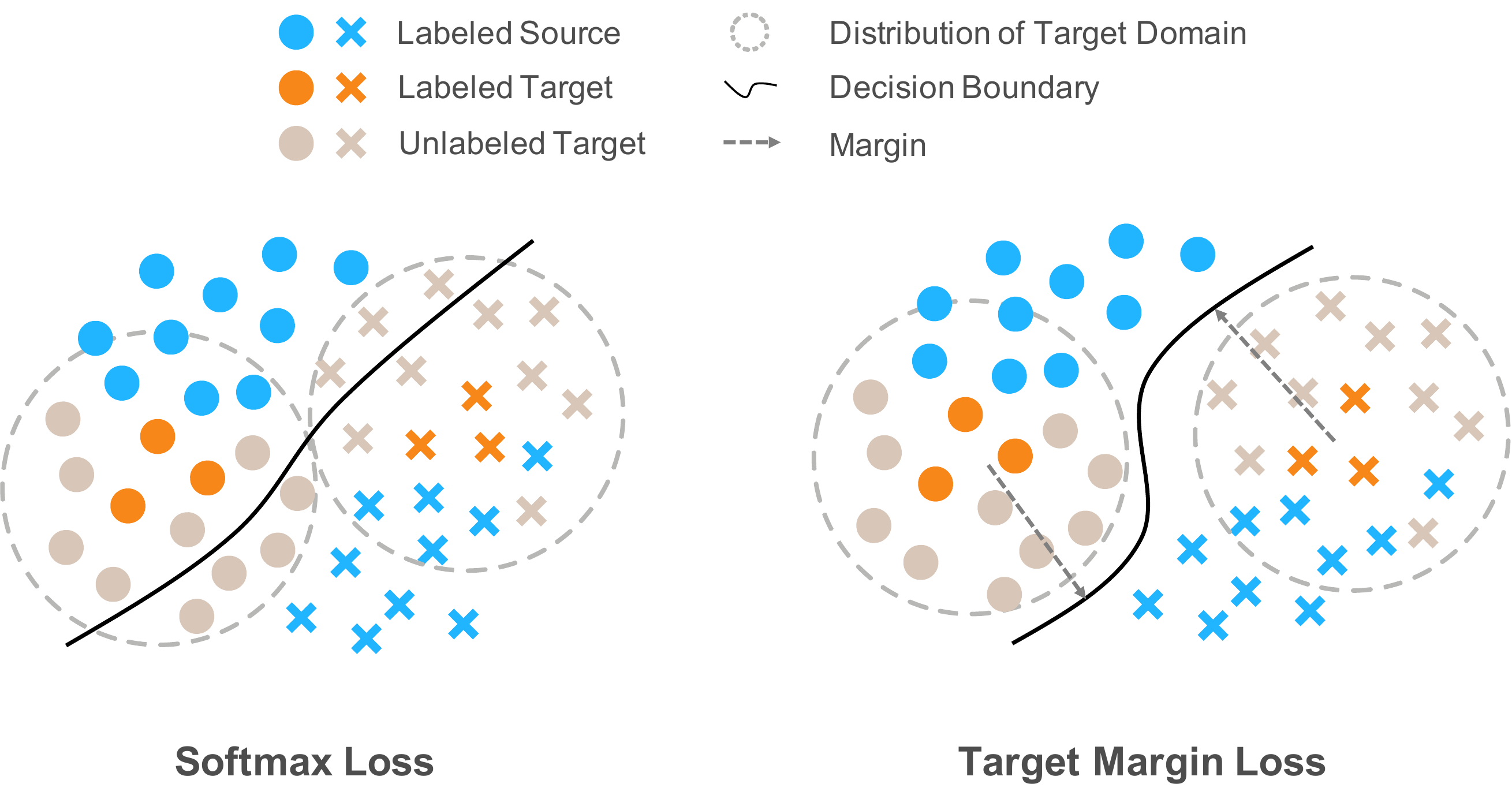}
\end{center}
   \caption{An illustration of the proposed target margin loss. In SSDA setting, the number of labeled target samples is much smaller than that of labeled source samples so that the decision boundary is mainly depending on the source domain. Our target margin loss can make the labeled target data more important for the decision boundary so that the decision regions in target domain will be more separated}
\label{fig:tml}
\end{figure}

In addition, due to the lack of labeled data in the target domain, the training methods typically have limited capacity in learning discriminative decision boundaries for the target domain. Inspired by the observation that large margin loss functions~\cite{liu2017sphereface,wang2018cosface,deng2019arcface} can help learn discriminative features, we propose a contrastive target margin loss function over the labeled data from the source and target domains . As illustrated in Fig.~\ref{fig:tml}, the target margin loss make the labeled target samples play a bigger role in learning the decision boundaries.

Overall, the contribution of this paper can be summarized as follows: (1) We propose a novel reinforcement learning framework for selective pseudo-labeling for semi-supervised domain adaptation. (2) We propose a contrastive target margin loss for SSDA that takes both generalization and discrimination into consideration. (3) Extensive experiments are conducted on DomainNet~\cite{peng2019moment}, Office-31~\cite{saenko2010adapting} and Office-Home~\cite{venkateswara2017deep}. The results show that our proposed method achieves the state-of-the-art semi-supervised domain adaptation performance.

\section{Related Work}

\paragraph{Domain Adaptation.}
Domain adaptation~\cite{wang2018deepda,gong2012geodesic,ganin2014unsupervised,sun2016return} aims to generalize models across different domains of different distributions. Most recent methods are focusing on unsupervised domain adaptation (UDA) which has a label-rich source domain and an unlabeled target domain. A widely used way in UDA methods is to add a domain discriminator which classifies whether a sample is drawn from the source or target domain. Then adversarial learning is applied to minimize the distance between the feature distributions in source and target domain. The domain-adversarial neural network (DANN)~\cite{ganin2014unsupervised} proposes the standard domain adversarial architecture and introduces a gradient reversal layer (GRL) for domain confusion loss produced by the discriminator. Inspired by Conditional Generative Adversarial Networks (CGANs)~\cite{goodfellow2014generative}, Long et al.~\cite{long2018conditional} proposes Conditional Domain Adversarial Network (CDAN) by combining the discriminative information conveyed in the classifier predictions into the adversarial adaptation. There are also other methods~\cite{saito2017adversarial,saito2018maximum} encouraging to generate more discriminative features for the target domain. In fact, semi-supervised domain adaptation (SSDA) is a more common setting in real-world datasets in which limited labeled data can be available in target domain. However, it has not been studied extensively, especially in the field of deep learning. A little conventional work~\cite{donahue2013semi,yao2015semi,ao2017fast} has concentrated on this important task. As for deep learning based methods, Saito et al.~\cite{saito2019semi} propose a Minimax Entropy (MME) method by alternately maximizing the conditional entropy of unlabeled target data and minimizing it to optimize the classifier and the feature extractor respectively. Meanwhile, ~\cite{saito2019semi} shows that UDA methods can rarely improve accuracy in SSDA. Therefore, we propose a reinforcement learning based pseudo-labeling method which can effectively exploit the information of target labeled data to generate more discriminative features with distributions closer to the target domain.

\paragraph{Pseudo-Labeling.}
Pseudo-labeling is an effective way to extend label set when the number of labels is limited. As for SSDA, pseudo-labeling can be used for target domain which has little labeled data. There are two strategies for pseudo-labeling without selection, hard labeling~\cite{long2013transfer,zhang2017joint,wang2018visual} and soft labeling~\cite{pei2018multi}. The hard labeling strategy assigns a pseudo-label with only one class predicted by the classifier to each target unlabeled instance. Then the pseudo-labeled target data will be combined with original labeled data to train an improved model. However, due to the weak classifier in the initial stage of training, many samples will be mis-labeled. Using these mis-labeled data for supervised training can cause serious harm to the model. Thus, soft labeling has been employed. Taking account of the confidence of prediction, the soft labeling strategy assigns the conditional probability of each class predicted by the classifier to the target unlabeled data. As for selective pseudo-labeling~\cite{zhang2018collaborative,wang2019unifying,chen2019progressive}, a subset of target unlabeled samples which are most confident in the prediction are selected to be pseudo-labeled. This sample selection strategy can make the pseudo-labels for training more accurate while it also has a limitation that the selected samples are generally too ``easy'' to represent the distribution of target data. In this work we use reinforcement learning to learn appropriate policies for selecting more accurate and representative pseudo-labeled samples.

\paragraph{Reinforcement Learning.}
Reinforcement learning (RL) is a technique which trains an agent to learn policies based on trial and error in a dynamic environment. The training strategy is to maximize the accumulated reward from the environment. No longer limited to its traditional applications in robotics and control, RL has made great progress in many vision tasks, such as action recognition~\cite{yeung2017learning}, person re-identification~\cite{zhang2018multi} and face recognition~\cite{liu2019fair}. Yeung et al.~\cite{yeung2017learning} design an agent for learning to label noisy web data so that they can select right examples for training a classifier. Zhang et al.~\cite{zhang2018multi} propose to choose sufficient data pairs for multi-shot pedestrian re-identification by training an agent. Liu et al.~\cite{liu2019fair} introduce a policy network for adjusting a margin parameter in the loss function to learn more discriminative features from imbalanced face datasets. In this work, we train an agent with deep Q-learning to select more representative and accurate pseudo-labeled data.

\begin{figure*}[htb]
\begin{center}
   \includegraphics[width=0.71\linewidth]{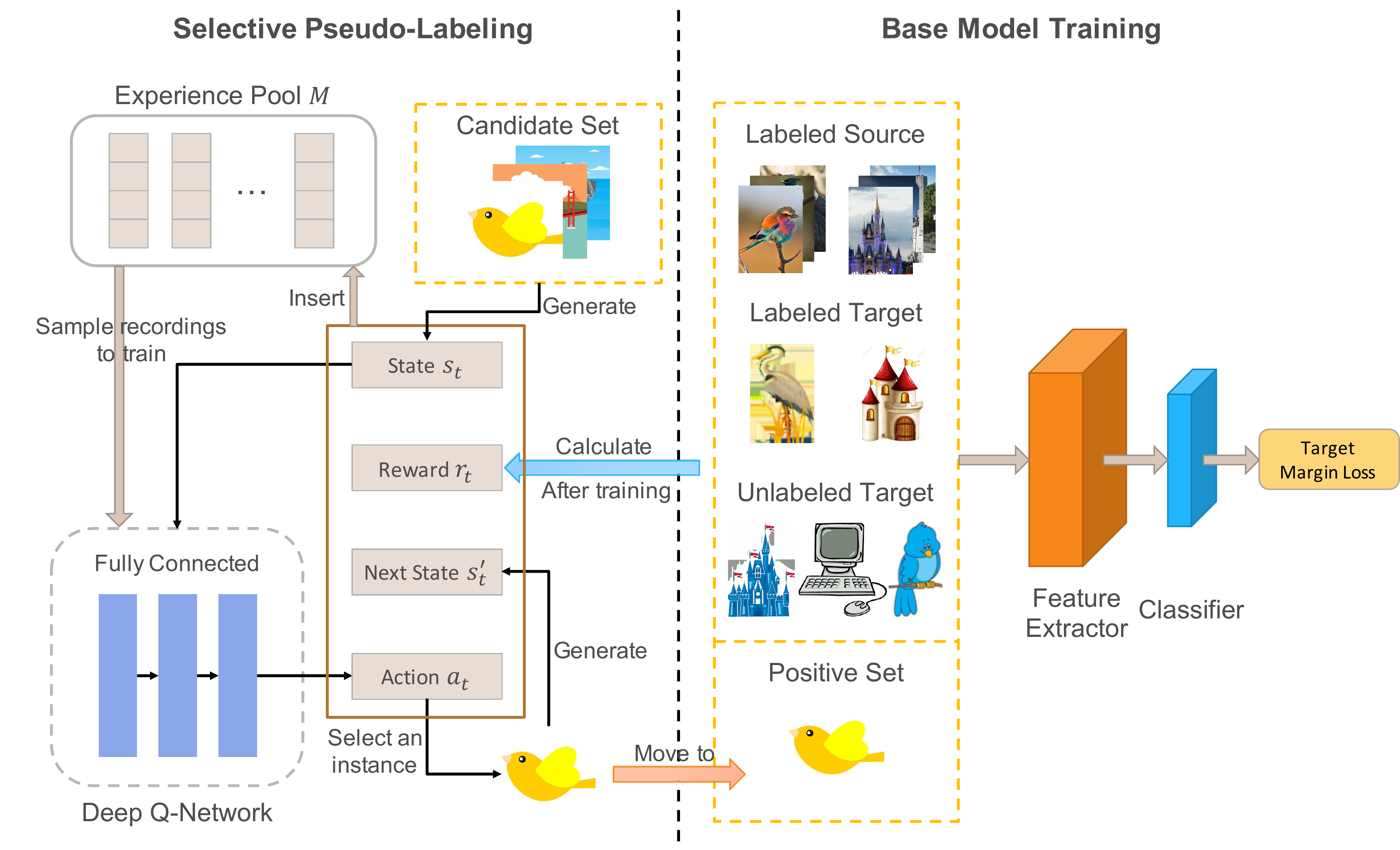}
\end{center}
   \caption{An overview of our method. We use the proposed target margin loss for the base model training. The candidate set consists of several samples pseudo-labeled by the pre-trained base model. Then we train an agent (deep Q-network) by deep Q-learning to select appropriate pseudo-labeled samples and move them to positive set for the improved model training. Specifically, we generate state ${s_t}$ from the candidate set as the input of the agent and get an output action ${a_t}$ to select an pseudo-labeled instance. Then the state transfers to the next state ${s'_t}$ and the reward ${r_t}$ will be calculated after an epoch of the improved model training. As for the training of the agent, we use an experience pool $M$ to collect a series of recordings $\left\{ {\left( {{s_i},{a_i},{r_i},{s'_i}} \right)} \right\}$ defined in section~\ref{sec:dqn} and sample batches of recordings from $M$ to train the deep Q-network}
\label{fig:model}
\end{figure*}

\section{Method}

For semi-supervised domain adaptation, we have a sufficient labeled dataset from the source domain, ${{\cal D}_s} = \left\{ {\left( {{\bf{x}}_i^s,y_i^s} \right)} \right\}_{i = 1}^{{N_s}}$. In the target domain, we only have a limited number of labeled instances ${{\cal D}_t} = \left\{ {\left( {{\bf{x}}_i^t,y_i^t} \right)} \right\}_{i = 1}^{{N_t}}$, but a large set of unlabeled instances ${{\cal D}_u} = \left\{ {\left( {{\bf{x}}_i^u} \right)} \right\}_{i = 1}^{{N_u}}$. The goal is to train a good prediction model on all these available data ${{\cal D}_s}$, ${{\cal D}_t}$ and ${{\cal D}_u}$ and evaluate it on ${{\cal D}_u}$ with the hidden true labels, as described in~\cite{saito2019semi}. In this section, we present a novel selective pseudo-labeling method for semi-supervised domain adaptation. The method is based on a deep Q-learning framework and a target margin loss. The framework of the proposed method is depicted in Fig.~\ref{fig:model}. First, we use the proposed target margin loss to train a CNN consisting of a feature extractor $F$ and a classifier $C$ for a K-class classification problem. Then, we generate pseudo-labels for the unlabeled samples in the target domain based on the trained CNN classifier. Finally, we alternately train an agent with deep Q-learning and use the agent to select pseudo-labeled samples for the CNN training.

\subsection{Target Margin Loss}

Large margin loss functions~\cite{liu2017sphereface,wang2018cosface,deng2019arcface} (based on traditional softmax loss function) effectively make CNN features more discriminative. For semi-supervised domain adaptation, there is a gap in feature distributions between domains and the number of labeled samples in the target domain is much smaller than in the source domain. Therefore, we propose to add a relative angular margin to the loss on the target labeled data, by contrast to the loss on the source labeled data. This can be considered as making the decision region separations more aligned with the target domain feature distribution. Then the proposed target margin loss can be formulated as follows:
\begin{equation}
\label{eq:tml}
\begin{aligned}
{{\cal L}_{tml}} = & - \frac{1}{{{N_s}}}\sum\limits_{i = 1}^{{N_s}} {\log \frac{{{e^{s\cos \theta _{{y_i}}^s}}}}{{\sum\nolimits_{j = 1}^K {{e^{s\cos \theta _j^s}}} }}} \\
& - \frac{1}{{{N_t}}}\sum\limits_{i = 1}^{{N_t}} {\log \frac{{{e^{s\cos \left( {\theta _{{y_i}}^t + m} \right)}}}}{{{e^{s\cos \left( {\theta _{{y_i}}^t + m} \right)}} + \sum\nolimits_{j = 1,j \ne {y_i}}^K {{e^{s\cos \theta _j^t}}} }}}
\end{aligned}
\end{equation}
where $\cos {\theta _j} = \frac{{{\bf{W}}_j^T}}{{\left\| {{{\bf{W}}_j}} \right\|}} \cdot \frac{{{F({\bf{x}}_i)}}}{{\left\| {{F({\bf{x}}_i)}} \right\|}}$ is the cosine similarity between the $i$-th instance feature vector and the $j$-th class weight. The $\cos ({\theta _{{y_i}}} + m)$ can be computed by the addition formula of the trigonometric functions with the values of $\cos {\theta _{{y_i}}}$ and the margin $m$. $s$ is a re-scaling constant. 

In addition, the suitable class separation can benefit from taking the unlabeled target domain data into account. To this end, we adopt the entropy loss to cluster the target unlabeled features into the corresponding decision regions, which is written as follows:
\begin{equation}
\label{eq:ent}
{{\cal L}_{ent}} =  - \frac{1}{{{N_u}}}\sum\limits_{i = 1}^{{N_u}} {\sum\limits_{j = 1}^K {p\left( {y = j|{\bf{x}}_i^u} \right)} } \log p\left( {y = j|{\bf{x}}_i^u} \right)
\end{equation}
where $p\left( {y = j|{\bf{x}}_i^u} \right)$ represents the probability of categorizing ${\bf{x}}_i^u$ to class $j$. Note that the samples should be clustered around the representative centers of the corresponding classes to decrease the entropy, resulting in the desired discriminative features.

Combining Eq. (\ref{eq:tml}) and Eq. (\ref{eq:ent}) provides the following formulation of our final semi-supervised loss function:
\begin{equation}
\label{eq:l}
{\cal L} = {{\cal L}_{tml}} + \alpha {{\cal L}_{ent}}
\end{equation}
where $\alpha$ is a hyper-parameter that balances the target margin loss and the entropy loss.

\subsection{Selective Pseudo-Labeling by Reinforcement Learning}
\label{sec:dqn}

Pseudo-labeling target unlabeled data by the CNN can generate many mis-labeled samples which can cause serious harm to the subsequent learning process. To address this issue, selective pseudo-labeling is used in many works, which selects the most confident unlabeled samples for pseudo-labeling. This sample selection strategy however is subject to the problem that the selected samples are generally ``easy'' ones or belong to ``easy'' classes. As a result, the selected samples could not represent the distribution of the target domain well. To address this drawback, we propose to use reinforcement learning to select more representative and accurate pseudo-labeled samples.

We formulate the problem of selecting pseudo-labeled samples as a Markov Decision Process (MDP), described by $\left( {{\cal S},{\cal A},{\cal T},{\cal R}} \right)$ as the states, actions, transitions and rewards respectively, and train an agent to select pseudo-labeled samples. We define a candidate set ${{\cal D}_c}$ which consists of pseudo-labeled samples to be selected and is initially randomly sampled from ${{\cal D}_u}$ with pseudo-labels, and a positive set ${{\cal D}_p}$ which is composed of the selected pseudo-labeled samples and initialized to be empty. During our CNN training, a series of reinforcement learning samples will be generated, which can be represented as $\left\{ {\left( {{s_i},{a_i},{r_i},{s'_i}} \right)} \right\}$. Here, ${s_i} \in {\cal S}$ is the state and ${r_i} \in {\cal R}$ is the reward. ${a_i} \in {\cal A}$ is the action taken by the agent at state ${s_i}$, which is equivalent to selecting a pseudo-labeled sample from the candidate set ${{\cal D}_c}$ and moving it to the positive set ${{\cal D}_p}$. ${s'_i} \in {\cal S}$ represents the next state which the agent turns to through the action ${a_i}$. Then, we alternately train the agent by using the reinforcement learning samples and use the agent to select pseudo-labeled samples for our CNN training.

\paragraph{States.} We consider that the representative ability and accuracy of a pseudo-labeled target instance can be related to three parts, which are itself, the data with labels in ${{\cal D}_t}$ and ${{\cal D}_p}$, and the unlabeled data in ${{\cal D}_u}$. Note that pseudo-labeled instances are selected to make the current distribution of labeled data similar to the distribution of unlabeled data. Therefore, we formulate the state as a concatenation of three vectors, dependent on ${{\cal D}_c}$, ${{\cal D}_t} \cup {{\cal D}_p}$ and ${{\cal D}_u}$, respectively. For the first part, given the candidate set ${{\cal D}_c} = \left\{ {\left( {{\bf{x}}_i^c,\hat y_i^c} \right)} \right\}_{i = 1}^{{N_c}}$, we use a vector $\left[ {F{{({\bf{x}}_i^c)}^T},C{{({\bf{x}}_i^c)}^T}} \right] \in {\mathbb{R}^{d + K}}$ to represent each instance, where $F({{\bf{x}}_i^c})$ denotes the $d$-dimensional feature vector of instance ${\bf{x}}_i^c$ extracted by the feature extractor $F$, $C({{\bf{x}}_i^c})$ denotes the softmax output of the classifier $C$. The entire vector of this part is a flattened concatenation of all the instances in ${{\cal D}_c}$. After taking an action to move an instance from ${{\cal D}_c}$ to ${{\cal D}_p}$, we replace the selected instance with a zero-valued vector. For the second part, we also use a vector $\left[ {F{{({\bf{x}}_i^{tp})}^T},C{{({\bf{x}}_i^{tp})}^T}} \right] \in {\mathbb{R}^{d + K}}$ to represent each instance in ${{\cal D}_t} \cup {{\cal D}_p}$. The difference is that we calculate the average vectors in each class and then concatenate them to a vector which has a dimension of $K \times \left( {d + K} \right)$. The last part is a vector represented by the instances in ${{\cal D}_u}$ with the same operation as the second part. Thus, the state ${s_i}$ is a flattened concatenation of these three parts.

\paragraph{Actions.} We define the action as selecting one instance from the current candidate set ${{\cal D}_c}$. For each state ${s_i}$, the agent takes an action ${a_i}$ to select the ${a_i}$-th instance in ${{\cal D}_c}$ and move it to ${{\cal D}_p}$. The number of actions is equivalent to the number of instances in ${{\cal D}_c}$, i.e., ${N_c}$.

\paragraph{Rewards.} The rewards should reflect whether the actions taken by the agent are appropriate or not. In other words, selecting more representative and accurate pseudo-labeled instances should lead to positive rewards, and vice versa. We first define a metric function to measure the representative ability and accuracy, which can be formulated as follows:
\begin{equation}
\label{eq:phi}
\varphi ({{\bf{x}}_i},{\hat y_i}) = \log {p_c}(y = {\hat y_i}|{{\bf{x}}_i}) + \beta \log {p_f}(y = {\hat y_i}|{{\bf{x}}_i}) + \lambda {\Delta _e}
\end{equation}
where $\beta$ and $\lambda$ are hyper-parameters. ${p_c}(y = {\hat y_i}|{{\bf{x}}_i})$ represents the probability of the pseudo class ${\hat y_i}$ predicted by the classifier. It indicates the confidence of the prediction. As for ${p_f}(y = {\hat y_i}|{{\bf{x}}_i})$, we take advantage of the target labeled data and define it as follows:
\begin{equation}
{p_f}(y = {\hat y_i}|{{\bf{x}}_i}) = \frac{{{e^{s\cos \left\langle {F({{\bf{x}}_i}),{\bf{z}}_{{{\hat y}_i}}^{tp}} \right\rangle }}}}{{\sum\nolimits_{j = 1}^K {{e^{s\cos \left\langle {F({{\bf{x}}_i}),{\bf{z}}_j^{tp}} \right\rangle }}} }}
\end{equation}
where ${\bf{z}}_j^{tp}$ represents the feature center of the $j$-th class in target labeled and current positive set ${{\cal D}_t} \cup {{\cal D}_p}$, which can be formulated as follows:
\begin{equation}
{\bf{z}}_j^{tp} = \frac{{\sum\nolimits_{i = 1}^{{N_t}} {F({{\bf{x}}_i})\mathbb{I}({y_i} = j) + \sum\nolimits_{i = 1}^{{N_p}} {F({{\bf{x}}_i})\mathbb{I}(\hat y_i^p = j)} } }}{{\sum\nolimits_{i = 1}^{{N_t}} {\mathbb{I}({y_i} = j) + \sum\nolimits_{i = 1}^{{N_p}} {\mathbb{I}(\hat y_i^p = j)} } }}
\end{equation}
$\mathbb{I}$ is the indicator function. Therefore, ${p_f}(y = {\hat y_i}|{{\bf{x}}_i})$ is the softmax output of the cosine distance between ${{\bf{x}}_i}$ and the feature center of its pseudo class ${\hat y_i}$ in ${{\cal D}_t} \cup {{\cal D}_p}$.

The first two terms reflect the confidence of the pseudo-label prediction through two aspects, i.e., the output of the classifier and the similarity with the feature center of the pseudo class in target domain. Since the classifier is more dependent on the source domain data, we add the second term to specifically take the distribution of the target domain into consideration so that the metric function can evaluate the accuracy of a target pseudo-labeled sample better. We also add a third term ${\Delta _e}$, which represents the decrease of the entropy of the target unlabeled data and can be formulated as follows:
\begin{equation}
{\Delta _e} = H - H'
\end{equation}
where $H$ and $H'$ represent the entropy at state ${s_i}$ and the next state ${s'_i}$ respectively and can be calculated in the same way asthe ${{\cal L}_{ent}}$ in Eq. (\ref{eq:ent}). In other words, we first calculate $H$ at state ${s_i}$ and then add a pseudo-labeled sample according to the action ${a_i}$ for one-epoch training. After the training, we calculate $H'$ and derive ${\Delta _e}$. In order to make the entropy not affected by ${{\cal L}_{ent}}$, we only use ${{\cal L}_{tml}}$ in Eq. (\ref{eq:tml}) to optimize the model during the one-epoch training. Note that the more representative the selected sample is, the more the entropy will decrease. Therefore, larger ${\Delta _e}$ means stronger representative ability, and vice versa. In addition, we perform a $\log$ operation on the first two terms to keep these three terms at the same scale.

Directly using the metric function as reward can result in very small differences between the rewards of good and bad actions. Hence, we propose to use the final reward function defined as follows:
\begin{equation}
\label{eq:rew}
{r_i} = \left\{ {\begin{array}{*{20}{c}}
{ + 1,\; \varphi ({{\bf{x}}_i},{{\hat y}_i}) > \tau }\\
{ - 1,\; \varphi ({{\bf{x}}_i},{{\hat y}_i}) \le \tau }
\end{array}} \right.
\end{equation}
where $\tau$ is a threshold and we set it to $(1 + \beta )\log (0.9)$ in our experiments. We use this binary reward instead of the metric function in order to provide the agent more explicit guidance. 

\paragraph{Deep Q-learning.} We apply deep Q-learning~\cite{mnih2015human} to learn policies for selecting pseudo-labeled instances. For each state and action $({s_i},{a_i})$, the Q function $Q({s_i},{a_i})$ can represent the discounted accumulated rewards for the state and action. Given a reinforcement learning training sample $\left( {{s_i},{a_i},{r_i},{s'_i}} \right)$, the target value of $Q({s_i},{a_i})$ can be calculated as follows:
\begin{equation}
\label{eq:v}
{V_i} = {r_i} + \gamma \mathop {\max }\limits_{{{a'}_i}} Q({s'_i},{a'_i})
\end{equation}
where $\gamma $ is a discount factor to decide the importance of future accumulated reward compared with the current reward. During the training, we iteratively update the deep Q-network by:
\begin{equation}
\label{eq:ql}
\Omega  \leftarrow \Omega  - \varepsilon \sum\limits_i {\frac{{{\rm{d}}Q({s_i},{a_i})}}{{{\rm{d}}\Omega }}(Q({s_i},{a_i}) - {V_i})} 
\end{equation}
where $\Omega $ represents the parameters of the Q-network. As for the entire model training, we alternately train the classification network and the Q-network. The details of our training strategy are summarized in Algorithm \ref{alg:dqn}.

\begin{algorithm}[htb]
\caption{Selective pseudo-labeling by deep Q-learning}
\label{alg:dqn}
\begin{algorithmic}[1]
\renewcommand{\algorithmicrequire}{\textbf{Input:}}
\REQUIRE
Source labeled set ${{\cal D}_s}$, target labeled set ${{\cal D}_t}$ and target unlabeled set ${{\cal D}_u}$
\renewcommand{\algorithmicrequire}{\textbf{Output:}}
\REQUIRE
Feature extractor $F$, classifier $C$ and deep Q-network $Q$
\STATE Pre-train $F$ and $C$ with ${{\cal D}_s}$, ${{\cal D}_t}$ and ${{\cal D}_u}$ by optimizing the loss in Eq. (\ref{eq:l});
\STATE Initialize the positive set ${{\cal D}_p} = \emptyset $;
\STATE Initialize the experience pool $M = \emptyset $;
\WHILE{not converge}
   \STATE Assign pseudo-labels to the unlabeled data in ${{\cal D}_u}$ by current $F$ and $C$;
   \STATE Copy the parameters of $F$ and $C$ to $F'$ and $C'$;
   \STATE Initialize ${{\cal D}_c}$ with random pseudo-labeled samples from ${{\cal D}_u}$ and generate the state ${s_0}$;
   \WHILE{${{\cal D}_c} \ne \emptyset $}
      \STATE Get an output action ${a_t}$ using Eq. (\ref{eq:act}) with the current state ${s_t}$ as input;
      \STATE Update ${{\cal D}_c}$ and ${{\cal D}_p}$ by taking the action ${a_t}$;
      \STATE Update $F'$ and $C'$ with ${{\cal D}_s}$, ${{\cal D}_t}$ and ${{\cal D}_p}$ by optimizing the loss in Eq. (\ref{eq:tml});
      \STATE Generate the next state ${s'_t}$;
      \STATE Calculate the reward ${r_t}$ by Eq. (\ref{eq:rew}) with $F'$ and $C'$;
      \STATE Insert the recording $\left( {{s_t},{a_t},{r_t},{s'_t}} \right)$ into $M$;
      \STATE Sample a batch of recordings $\left\{ {\left( {{s_i},{a_i},{r_i},{s'_i}} \right)} \right\}$ from $M$ to update the deep Q-network $Q$ by Eq. (\ref{eq:ql});
      \IF{${r_t} < 0$}
         \STATE Break;
      \ENDIF
   \ENDWHILE
   \STATE Update $F$ and $C$ with ${{\cal D}_s}$, ${{\cal D}_t}$, ${{\cal D}_u}$ and ${{\cal D}_p}$ by optimizing the loss in Eq. (\ref{eq:l}).
\ENDWHILE
\end{algorithmic}
\end{algorithm}

We simply use a three-layer fully connected network as the deep Q-network. Each fully connected layer is followed by a ReLU activation function. When the agent (i.e., the deep Q-network) is called to select a pseudo-labeled instance, it will output an action ${a_t}$ using a policy as follows:
\begin{equation}
\label{eq:act}
{a_t} = \arg {\max _a} \; Q({s_t},a)
\end{equation}
where ${s_t}$ is the current state representation.

\section{Experiments}

\subsection{Datasets and Baselines}

\paragraph{Datasets.}
We perform our experiments on three datasets, DomainNet~\cite{peng2019moment}, Office-31~\cite{saenko2010adapting} and Office-Home~\cite{venkateswara2017deep}. DomainNet is a large-scale domain adaptation benchmark dataset, which contains six domains. Compared with the other two datasets, it has much more data with 345 classes. Office-31 is a widely used dataset for visual domain adaptation, with 31 categories collected from three different domains: Amazon (A), Webcam (W) and DSLR (D). Office-Home is another domain adaptation dataset in office and home settings, which is more difficult than Office-31. It consists of four distinct domains Artistic images (A), Clipart (C), Product images (P) and Real-World images (R). For the SSDA setting and domain adaptation scenarios, we follow the strategies in ~\cite{saito2019semi}. We randomly select three labeled samples per class as the labeled training target samples to form a three-shot SSDA setting. Due to some noisy domains and categories in DomainNet, we pick 4 domains and 126 categories for DomainNet experiments. Following ~\cite{saito2019semi}, we form 7 adaptation scenarios for testing with the 4 domains, Real (R), Clipart (C), Painting (P) and Sketch (S). As for Office-31, we construct 2 scenarios with Amazon as the target domain since Webcam and DSLR do not have enough samples for effective evaluation.

\paragraph{Baselines.}
S+T baseline directly trains a model using the labeled
source data and labeled target data without unlabeled target data. For the UDA methods (DANN~\cite{ganin2014unsupervised}, ADR~\cite{saito2017adversarial}, CDAN~\cite{long2018conditional}) as baselines, we modify their training strategies following ~\cite{saito2019semi} so that the models can be trained with all the labeled source set, labeled target set and unlabeled target set. ENT~\cite{grandvalet2005semi} is a baseline method using standard entropy minimization which trains a model to minimize the entropy of unlabeled target data. MME~\cite{saito2019semi} is a model with the classifier and the feature extractor optimized by maximizing the conditional entropy of unlabeled target data and minimizing it respectively. We also design a baseline TML\_SPL with a selective pseudo-labeling strategy to compare with our reinforcement learning based selective pseudo-labeling method. TML\_SPL uses our target margin loss for the entire training and selects the most confident pseudo-labeled samples with 0.9 as the threshold to assist to train an improved model.

\subsection{Implementation Details}

As for the architecture of the networks, we adopt ResNet-34~\cite{he2016deep} for experiments on DomainNet and VGG-16~\cite{simonyan2014very} for experiments on Office-31 and Office-Home, finetuned from ImageNet pre-trained models~\cite{russakovsky2015imagenet}. All our experiments are implemented in PyTorch~\cite{paszke2017automatic}. We adopt mini-batch SGD with momentum of 0.9 and the learning rate adjusting schedule as ~\cite{ganin2016domain}. The weight decay is set at 0.0005. As for the $s$ and margin $m$ in Eq. (\ref{eq:tml}), a very small $s$ or a very large $m$ can make the model difficult to converge while a very large $s$ or a very small $m$ can make the margin ignored during training. Therefore, we choose an appropriate pair of values 30 and 0.5 for $s$ and $m$ respectively following previous large margin works. As for the trade-off parameters, we set $\alpha$ in Eq. (\ref{eq:l}) at 0.1 to keep the two losses at a similar scale so that we can prevent any loss from being ignored during the training. $\beta$ and $\lambda$ balance the second and the third terms in Eq. (\ref{eq:phi}). The first two terms in Eq. (\ref{eq:phi}) are both logarithmic forms and the third term is the decrease of the entropy. Thus, we set $\beta$ and $\lambda$ at 1 and 0.1 to keep the three terms at a similar scale so that all the terms can make sense.

For the deep Q-learning, we apply the $\varepsilon $-greedy strategy~\cite{mnih2015human} and the experience replay strategy~\cite{lin1992self}. The $\varepsilon $-greedy strategy allows the deep Q-network to randomly select an action with $\varepsilon $ as the probability and $\varepsilon $ changes from 1 to 0 during the training. We use this strategy for the reason that the output by the deep Q-network at early stage does not reflect the reward and the deep Q-network needs more diverse training samples. The experience replay strategy means that we use an experience pool $M$ to collect a series of recordings $\left\{ {\left( {{s_i},{a_i},{r_i},{s'_i}} \right)} \right\}$ and sample batches of the recordings from $M$ to train the deep Q-network. It can make the deep Q-network learn from both current and past information. Our deep Q-network is composed of three fully connected layers, with two hidden layers of 1024 and 512 units respectively. The discount factor $\gamma $ in Eq. (\ref{eq:v}) is set to 0.9.

\setlength{\tabcolsep}{2pt}
\begin{table}[htb]
\begin{center}
\caption{Comparisons for CML and TML using DomainNet based on ResNet-34}
\scalebox{0.8}{
\begin{tabular}{c|ccccccc|c}
\hline
Method&R to C&R to P&P to C&C to S&S to P&R to S&P to R&Mean\\
\hline\hline
CML&66.3&63.8&67.2&58.8&61.0&57.2&71.7&63.7\\
TML (Ours)&\textbf{72.5}&\textbf{71.6}&\textbf{72.9}&\textbf{61.0}&\textbf{67.7}&\textbf{62.8}&\textbf{79.2}&\textbf{69.8}\\
\hline
\end{tabular}
}
\label{tab:cml}
\end{center}
\end{table}
\setlength{\tabcolsep}{5pt}

\begin{figure}[htb]
\begin{center}
   \includegraphics[width=0.93\columnwidth]{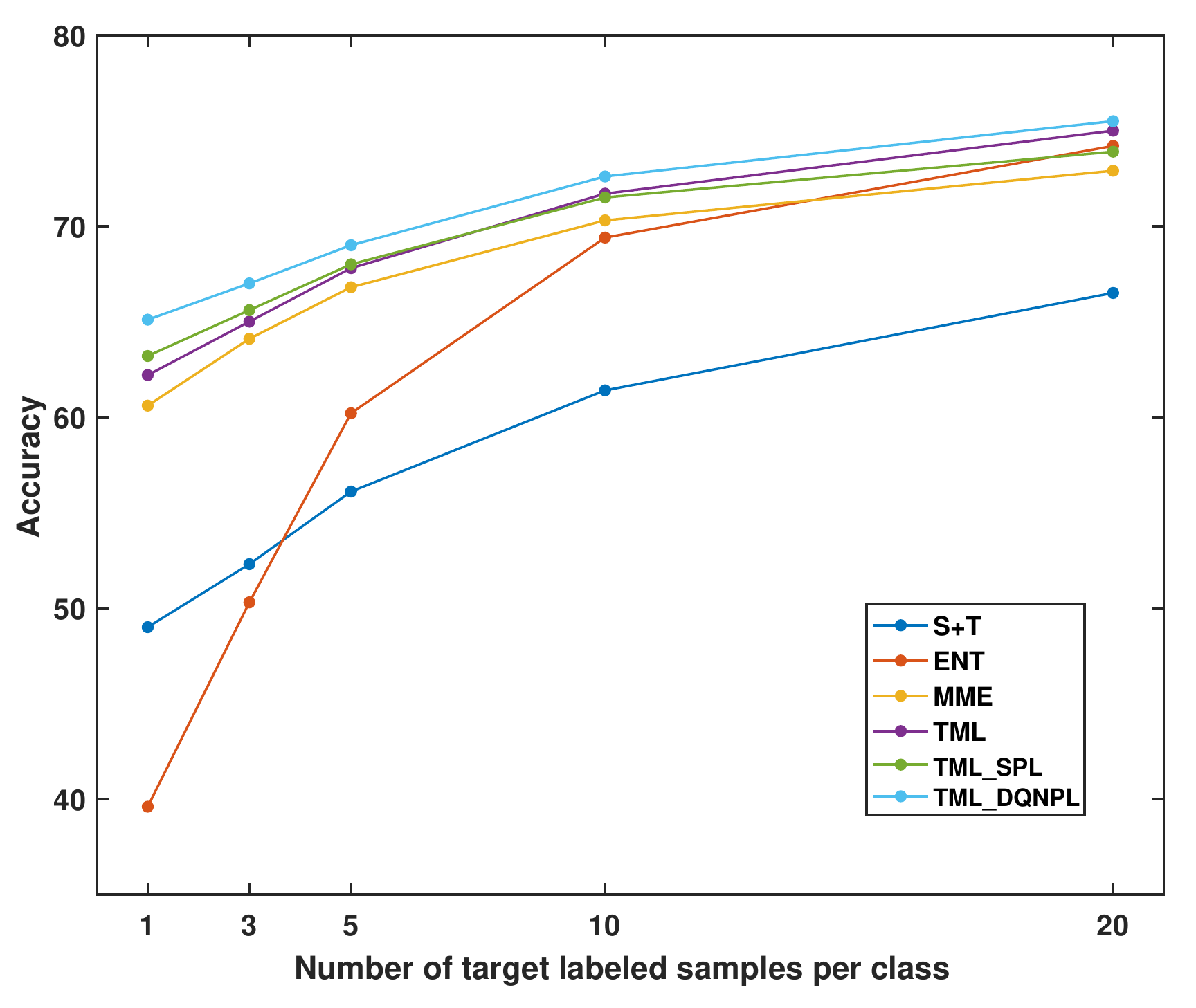}
\end{center}
\vspace{-2ex}
   \caption{Accuracy vs the number of labeled samples per class in target domain}
\label{fig:vn}
\vspace{-1ex}
\end{figure}

\subsection{Validation Experiments}

In order to show the effectiveness of our method, we design several validation experiments. We pick a scenario in DomainNet as an example scenario with Real as the source domain and Clipart as the target domain.

\paragraph{Extending Margin to Source Domain.}
Margin loss can make features more discriminative while our target margin loss only adds a margin on target domain. We extend the margin parameter to source domain to form a complete margin loss (CML) for comparison, which has a similar formulation to the original margin loss and can be written as:
\begin{equation}
{{\cal L}_{cml}} = \frac{1}{{{N_{s + t}}}}\sum\limits_{i = 1}^{{N_{s + t}}} {\log \frac{{{e^{s\cos \left( {\theta _{{y_i}}^t + m} \right)}}}}{{{e^{s\cos \left( {\theta _{{y_i}}^t + m} \right)}} + \sum\nolimits_{j = 1,j \ne {y_i}}^K {{e^{s\cos \theta _j^t}}} }}} 
\end{equation}
where the $s + t$ means all the data in labeled source set and labeled target set. The CML method replaces the ${{\cal L}_{tml}}$ in Eq. (\ref{eq:l}) with ${{\cal L}_{cml}}$. The results are shown in Table~\ref{tab:cml}. Our TML method performs much better than CML though the difference between the two methods is quite small. The reason can be that if both the labeled source data and the labeled target data is constrained by the margin then the decision boundary will still be more dependent on the source domain with much more data.

\begin{table*}[htb]
\begin{center}
\caption{Results on the 7 scenarios in DomainNet based on ResNet-34. The top-performing method in each scenario is bold and the best method without pseudo-labeling is underlined.}
\scalebox{0.9}{
\begin{tabular}{c|ccccccc|c}
\hline
Method&R to C&R to P&P to C&C to S&S to P&R to S&P to R&Mean\\
\hline\hline
S+T&60.0&62.2&59.4&55.0&59.5&50.1&73.9&60.0\\
DANN~\cite{ganin2014unsupervised}&59.8&62.8&59.6&55.4&59.9&54.9&72.2&60.7\\
ADR~\cite{saito2017adversarial}&60.7&61.9&60.7&54.4&59.9&51.1&74.2&60.4\\
CDAN~\cite{long2018conditional}&69.0&67.3&68.4&57.8&65.3&59.0&78.5&66.5\\
ENT~\cite{grandvalet2005semi}&71.0&69.2&71.1&60.0&62.1&61.1&78.6&67.6\\
MME~\cite{saito2019semi}&72.2&69.7&71.7&\underline{61.8}&66.8&61.9&78.5&68.9\\
TML (Ours)&\underline{72.5}&\underline{71.6}&\underline{72.9}&61.7&\underline{67.7}&\underline{62.8}&\underline{79.2}&\underline{69.8}\\
\hline
TML\_SPL&73.2&72.2&73.3&62.1&68.5&63.4&80.0&70.4\\
TML\_DQNPL (Ours)&\textbf{75.8}&\textbf{74.5}&\textbf{75.1}&\textbf{64.3}&\textbf{69.7}&\textbf{64.4}&\textbf{82.6}&\textbf{72.3}\\
\hline
\end{tabular}
}
\label{tab:domainnet}
\end{center}
\vspace{-1ex}
\end{table*}

\begin{table*}[htb]
\begin{center}
\caption{Results on the 12 scenarios in Office-Home based on VGG-16. The top-performing method in each scenario is bold and the best methods without pseudo-labeling are underlined.}
\scalebox{0.88}{
\begin{tabular}{c|cccccccccccc|c}
\hline
Method&R to C&R to P&R to A&P to R&P to C&P to A&A to P&A to C&A to R&C to R&C to A&C to P&Mean\\
\hline\hline
S+T&49.6&78.6&63.6&72.7&47.2&55.9&69.4&47.5&73.4&69.7&56.2&70.4&62.9\\
DANN&56.1&77.9&63.7&73.6&52.4&56.3&69.5&50.0&72.3&68.7&56.4&69.8&63.9\\
ADR&49.0&78.1&62.8&73.6&47.8&55.8&69.9&49.3&73.3&69.3&56.3&71.4&63.0\\
CDAN&50.2&80.9&62.1&70.8&45.1&50.3&74.7&46.0&71.4&65.9&52.9&71.2&61.8\\
ENT&48.3&81.6&65.5&76.6&46.8&56.9&73.0&44.8&75.3&72.9&59.1&77.0&64.8\\
MME&\underline{56.9}&82.9&65.7&76.7&53.6&59.2&\underline{75.7}&\underline{54.9}&75.3&72.9&\underline{61.1}&76.3&67.6\\
TML (Ours)&\underline{56.9}&\underline{83.2}&\underline{67.0}&\underline{76.8}&\underline{54.5}&\underline{59.9}&\underline{75.7}&\underline{54.9}&\underline{75.9}&\underline{73.2}&\underline{61.1}&\underline{77.5}&\underline{68.1}\\
\hline
TML\_SPL&55.4&82.1&67.1&76.5&55.3&60.7&75.5&53.0&75.9&73.4&60.4&77.6&67.7\\
TML\_DQNPL (Ours)&\textbf{58.4}&\textbf{84.0}&\textbf{69.1}&\textbf{78.5}&\textbf{56.8}&\textbf{61.7}&\textbf{77.0}&\textbf{55.9}&\textbf{77.1}&\textbf{74.5}&\textbf{61.9}&\textbf{78.8}&\textbf{69.5}\\
\hline
\end{tabular}
}
\label{tab:officehome}
\end{center}
\vspace{-1ex}
\end{table*}

\begin{table}[htb]
\begin{center}
\caption{Results on the 2 scenarios in Office-31 based on VGG-16. The top-performing method in each scenario is bold and the best methods without pseudo-labeling are underlined.}
\scalebox{0.9}{
\begin{tabular}{c|cc|c}
\hline
Method&W to A&D to A&Mean\\
\hline\hline
S+T&73.2&73.3&73.3\\
DANN&75.4&74.6&75.0\\
ADR&73.3&74.1&73.7\\
CDAN&74.4&71.4&72.9\\
ENT&75.4&75.1&75.3\\
MME&76.3&\underline{77.6}&77.0\\
TML (Ours)&\underline{76.6}&\underline{77.6}&\underline{77.1}\\
\hline
TML\_SPL&75.7&77.2&76.5\\
TML\_DQNPL (Ours)&\textbf{77.5}&\textbf{78.8}&\textbf{78.2}\\
\hline
\end{tabular}
}
\label{tab:office}
\end{center}
\vspace{-1ex}
\end{table}

\paragraph{Varying Number of Target Labeled Samples.}
We verify the number of labeled samples in the target domain from 1 to 20 per class to explore the performance of our method in different settings.
As illustrated in Fig.~\ref{fig:vn}, our TML method can outperform MME and ENT in all the settings with varying number of target labeled samples while MME gradually performs worse than the simple ENT baseline as the number increasing. Furthermore, when the target labeled samples are much enough, the confidence based selective pseudo-labeling method TML\_SPL doesn't work well and can even hurt the original model. Our reinforcement learning based selective pseudo-labeling method TML\_DQNPL can always make progress to the base model due to the representative and accurate selected pseudo-labels.

\subsection{Results}

The results of our main experiments on the DomainNet dataset are shown in Table~\ref{tab:domainnet}. As mentioned in ~\cite{saito2019semi}, the UDA methods cannot improve the performance in some cases. Compared with the present state-of-the-art SSDA method MME~\cite{saito2019semi}, our method with only target margin loss (TML) can perform better except for only one case where it performs similarly to MME. On the basis of TML, our final method TML\_DQNPL with deep Q-network for selective pseudo-labeling can outperform the baseline TML\_SPL with selective pseudo-labeling by confidence, which demonstrates that our deep Q-network can help select more representative and accurate pseudo-labels.

The results on Office-Home and Office-31 are shown in Table~\ref{tab:officehome} and Table~\ref{tab:office} respectively. With these small-scale datasets, our TML method also has better performance than MME in most cases and can perform the same as MME in other cases. In addition, we observe that TML\_SPL can hurt the performance in some cases while our TML\_DQNPL also makes a progress. The potential reason can be that adding mis-labeled target samples for training causes more damage to the model when the number of the original training samples is small in small-scale datasets. Therefore, these comparisons can further confirm the effectiveness of our method.

\section{Conclusions}

We propose a novel reinforcement learning based selective pseudo-labeling method for semi-supervised domain adaptation (SSDA). We first design a target margin loss for our base model training, which can make the feature distribution closer to the target domain and improve the discriminative ability. Then we apply deep Q-learning to train an agent to select more representative and accurate pseudo-labeled samples for our improved model training. Our method obtains competitive results on several domain adaptation benchmarks and outperforms the present state-of-the-art methods. In addition, the training of the deep Q-network is unrelated to the architecture of base model and will not change the training strategy of base model so that the proposed pseudo-labeling agent can be combined with other advanced methods to help train improved models in the SSDA setting.

\bibliography{egbib}

\end{document}